\documentclass[10pt,twocolumn,letterpaper]{article}

\usepackage{iccv}
\usepackage{times}
\usepackage{epsfig}
\usepackage{graphicx}
\usepackage{amsmath}
\usepackage{amssymb}

\usepackage{enumitem}
\usepackage{amsmath,amssymb}
\usepackage{bbm}
\usepackage{multirow}
\usepackage[export]{adjustbox}
\usepackage{gensymb}
\usepackage{floatrow}
\usepackage{comment}
\usepackage{float}
\usepackage{xspace}
\usepackage{tabularray}
\UseTblrLibrary{booktabs}
\SetTblrInner{rowsep=1pt}
\usepackage[dvipsnames]{xcolor}
\usepackage{makecell}
\usepackage{cuted}

\newcommand{\ldpdesc}{\textsc{LDP-Feat}\xspace}

\usepackage{url}

\usepackage[pagebackref=true,breaklinks=true,letterpaper=true,colorlinks,bookmarks=false,citecolor=blue]{hyperref}

\usepackage[hyphenbreaks]{breakurl}

\usepackage[capitalize]{cleveref}
\crefname{section}{Sec.}{Secs.}
\Crefname{section}{Section}{Sections}
\Crefname{table}{Table}{Tables}
\crefname{table}{Tab.}{Tabs.}
\crefname{equation}{Eq.}{Eqs.}

\iccvfinalcopy 
\def\iccvPaperID{1478} 

\ificcvfinal\fi

\begin{document}

\title{\ldpdesc: Image Features with Local Differential Privacy}

\author{\quad Francesco Pittaluga \quad \quad \quad \quad \quad \quad \quad
Bingbing Zhuang\\
{\tt\small francescopittaluga@nec-labs.com} \quad \quad \quad {\tt\small bzhuang@nec-labs.com} \\
\quad NEC Labs America
}

\maketitle
\ificcvfinal\thispagestyle{empty}\fi


\begin{abstract}
Modern computer vision services often require users to share raw feature descriptors with an untrusted server. This presents an inherent privacy risk, as raw descriptors may be used to recover the source images from which they were extracted. To address this issue, researchers~\cite{dusmanu2021privacy} recently proposed privatizing image features by embedding them within an affine subspace containing the original feature as well as adversarial feature samples. In this paper, we propose two novel inversion attacks to show that it is possible to (approximately) recover the original image features from these embeddings, allowing us to recover privacy-critical image content. In light of such successes and the lack of theoretical privacy guarantees afforded by existing visual privacy methods, we further propose the first method to privatize image features via local differential privacy, which, unlike prior approaches, provides a guaranteed bound for privacy leakage regardless of the strength of the attacks. In addition, our method yields strong performance in visual localization as a downstream task while enjoying the privacy guarantee. 
\end{abstract}

\section{Introduction}

The extraction and matching of image keypoints with descriptors is an essential building block for many vision problems, such as 3D reconstruction~\cite{agarwal2011}, image retrieval~\cite{nister2006} and recognition~\cite{turk1991}. Thus, modern computer vision services often require the users to share feature descriptors and/or raw images to a centralized server for downstream tasks, such as visual localization~\cite{speciale2019}. However, recent works~\cite{pittaluga2019revealing,song2020deep} show that high-quality images may be recovered from the keypoint descriptors~\cite{pittaluga2019revealing} or their spatial information~\cite{song2020deep}, raising serious concerns regarding the potential leakage of private information via inversion attacks. 

\begin{figure}[]
  \centering
  \includegraphics[width=1.0\linewidth]{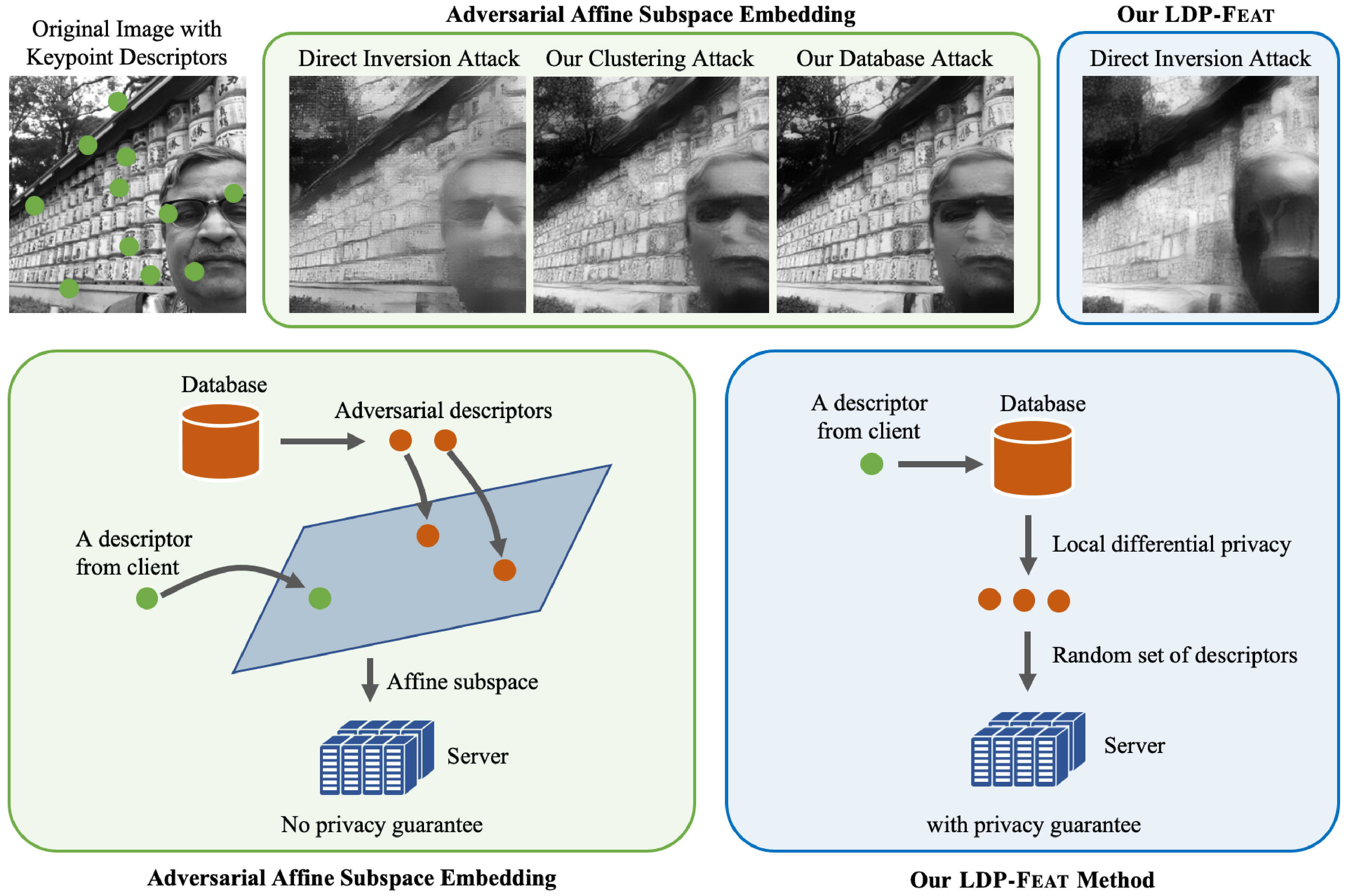}
  \centering
  \caption{\textbf{Our novel privacy attacks and LDP-based privacy method.} Top row: image reconstruction attacks against adversarial affine subspace embeddings \cite{dusmanu2021privacy} and our \ldpdesc when they have comparable performance in the downstream utility task. Bottom row: overviews of the adversarial affine subspace embeddings algorithm \cite{dusmanu2021privacy} and our \ldpdesc algorithm}
  \label{fig:teaser}
\end{figure}

This in turn inspires great interest in researching feature obfuscation with a view to concealing the privacy critical information in the image, mainly by perturbing either the feature descriptors or their locations.
One of the recent works~\cite{dusmanu2021privacy} represents a feature descriptor point as an affine subspace passing through the original point, as well as a number of other adversarial descriptors randomly sampled from a database of descriptors, as shown in \cref{fig:teaser}. These adversarial descriptors serve as confounders to conceal the raw descriptor. Another line of research~\cite{speciale2019,speciale2019privacy,geppert2020privacy,shibuya2020privacy} aims to conceal the location of 2D or 3D keypoints by lifting the point to a line passing through that point, which prevents a direct attack of the sort in~\cite{pittaluga2019revealing,song2020deep}. 

Despite their success, these works are primarily evaluated on the basis of empirical performance of a chosen attacker, without rigorous understanding of the attacker-independent, intrinsic privacy property. This causes hindrance for a method to claim privacy protection safely since there is no theoretical guarantee to assure practical applications. For instance, \cite{chelani2021} re-investigates the privacy claim in~\cite{speciale2019} and designs a stronger attack to reveal that a significant amount of scene geometry information in fact still exists in the lifted line clouds, which can be leveraged to recover sensitive image content. In this paper, we focus on the feature descriptor and, similar in spirit to~\cite{chelani2021}, we reveal the privacy leakage in the affine subspace lifting~\cite{dusmanu2021privacy}. Considering the drawbacks of the visual privacy-based method, we present the first attempt of its kind to formulate the privacy protection of image features through the lens of differential privacy~\cite{wang2017locally}, which permits theoretic privacy characterization, enjoys a guaranteed bound on privacy loss, 
and has become a gold standard notion of privacy.

More specifically, we firstly introduce two novel attacks against the adversarial affine subspace embedding~\cite{dusmanu2021privacy}, namely the \textit{database attack} and the \textit{clustering attack}. In the database attack, we assume that the database used to sample the adversarial descriptors is accessible to the attacker, while in clustering attack we relax this assumption. At its core, both attacks are established based upon the following key assumption: the low-dimensional (\eg 2,4,8) affine subspace very likely only intersects with the manifold of high-dimensional (\eg 128 for SIFT~\cite{lowe2004distinctive}) descriptors at those points that were intentionally selected to construct the subspace in the beginning, i.e. the raw descriptor 
 to be concealed and the adversarial ones chosen from the database. Tha main idea of our attacks lies in identifying these intersections and further eliminating the adversarial ones. As shown in \cref{fig:teaser}, our attacks recover images of higher quality than the direct inversion attack shown in \cite{dusmanu2021privacy}.

Next, we propose \ldpdesc, a novel descriptor privatization method that rests on the notion of local differential privacy (LDP)~\cite{yang2020local}, as illustrated in \cref{fig:teaser}. In contrast to the original centralized differential privacy which prevents private information in the database from releasing to queries, we instead aim to protect privacy in the query itself, \ie the image descriptors to be sent. We propose to formulate the feature obfuscation by local differential privacy, with the so-called $\omega$-subset mechanism~\cite{wang2017locally} -- we effectively replace each raw descriptor with a random set of descriptors under predefined probability distribution that endows the rigorous and quantifiable differential privacy guarantee. Further, our database and clustering attack are not applicable to \ldpdesc, and the direct inversion attack largely fails on \ldpdesc, as shown in \cref{fig:teaser}. We demonstrate strong performance in visual localization as a downstream task while enjoying the advantageous privacy guarantee. 

\noindent In summary, our contributions include:
\begin{itemize}[noitemsep,nolistsep]
\item  Two novel attacks on adversarial affine subspace embeddings~\cite{dusmanu2021privacy} that enable (approximate) recovery of the original feature descriptors.
\item A novel method for image feature privatization that rests on local differential privacy with favorable privacy guarantees.
\item Advantageous privacy-utility trade-offs achieved via empirical results to support practical applications.
\end{itemize}

\section{Related Work}

\noindent \textbf{Feature descriptors.~}
Feature descriptors extracted from image key points are used for a range of computer vision tasks such as 3D scene reconstruction \cite{agarwal2011}, image retrieval \cite{nister2006} and recognition~\cite{turk1991}. Traditional methods for extracting such descriptors were handcrafted based on direct pixel sampling \cite{calonder2011brief} or histograms of image gradients \cite{lowe2004distinctive,dalal2005histograms}. More recently, a growing number of methods rely on deep learning to extract the feature descriptors \cite{mishchuk2017working,he2018local,balntas2016learning}. 

\noindent \textbf{Inverting image features.~} 
The task of reconstructing images from features has been explored to understand what is encoded by the features, as was done for SIFT by~\cite{weinzaepfel2011reconstructing}, HOG features by~\cite{vondrick2013hoggles} and bag-of-words by~\cite{kato2014image}. Recent work has been primarily focused on inverting and interpreting CNN features~\cite{zeiler2014visualizing,yosinski2015understanding,mahendran2015understanding}. Dosovitskiy and Brox~\cite{dosovitskiy2016inverting} proposed encoder-decoder CNN architectures for
inverting many different features and later incorporated adversarial training with a perceptual loss~\cite{dosovitskiy2016generating}, primarily focusing on dense features. Pittaluga et al.~\cite{pittaluga2019revealing} focus on inverting sparse SIFT descriptors stored along with structure-from-motion point clouds, recovering high-quality image content by training feature inversion networks. Song et al.~\cite{song2020deep} further demonstrate image recovery from just colored 3D point clouds without the descriptors. The capability enabled by these works raise significant privacy concerns, which further motivate research in privacy-preserving visual representations. 

\noindent \textbf{Visual privacy.~}
McPherson et al.~\cite{mcpherson2016} and Vasiljevic et al.~\cite{vasiljevic2016examining} showed that deep models could defeat conventional image obfuscation methods such as blurring and pixelation.  To defend against CNN-based attacks, adversarial optimization has been employed to learn CNN-resistant image encodings for action recognition \cite{wu2018towards,wang2019privacy}, face attribute recognition \cite{xiao2020adversarial}, place recognition \cite{pittaluga2019learning},  and more \cite{cai2021generative}. Researchers also developed privacy-preserving representations for image-based localization and mapping, which is tackled from two angles: (i) 2D and 3D keypoint obfuscation \cite{speciale2019, speciale2019privacy, shibuya2020privacy, geppert2021privacy,chelani2021,geppert2020privacy,geppert2022privacy} by concealing the position information of keypoints, and (ii) image feature/descriptor obfuscation \cite{dusmanu2021privacy, ng2021ninjadesc} by concealing the descriptor of keypoints. The main idea for keypoint obfuscation lies in lifting a point to a random line or plane, while descriptor obfuscation lifts the descriptor to an affine subspace~\cite{dusmanu2021privacy}, or directly learns attack-resistant descriptors by adversarial training~\cite{ng2021ninjadesc}. Despite their empirical effectiveness, these methods do not provide theoretical guarantee and characterization on the privacy protection. In this paper, we first introduce two novel attacks against \cite{dusmanu2021privacy} that reveal its privacy leakage, and then propose a new feature privatization method that provides formal guarantees via local differential privacy.

\noindent \textbf{Differential privacy.~}
In recent years, differential privacy \cite{dwork2006differential,dwork2006calibrating} has become the gold standard for publication and analysis of sensitive data. Most differential privacy research is focused on the centralized setting \cite{hassan2019differential}, where raw user data is aggregated by a trusted curator who then shares the data to the public without releasing private information. Note, the curator is assumed trusted, which, however, may not be the case in practice. Instead, the local differential privacy~\cite{wang2017locally, yang2020local} provides a means for users to privatize their data locally prior to sending it out; hence a trusted curator is not required.  Our work presents the first attempt to apply local differential privacy for image feature privatization.

\section{Preliminary: Affine Subspace Embedding}
\label{sec:inv_aase}

In order to preserve privacy in keypoint descriptors, Dusmanu et al.~\cite{dusmanu2021privacy} propose to ``lifting'' each descriptor to an adversarial affine subspace before sharing to the curator. 

\noindent \textbf{Subspace lifting.}  Let $d \in \mathbb{R}^n$ denote a descriptor to be privatized. \cite{dusmanu2021privacy} proposes lifting $d$ to an $m$-dimensional affine subspace $D \subset \mathbb{R}^n$ satisfying $d \in D$, represented by a translation vector $d_0$ and $m$ basis vectors $\{d_1,...,d_m\}$, i.e. $D=d_0+\mathtt{span}(d_1,...d_m)$.

\noindent \textbf{Selection of subspace.} To ensure $d$ not be easily recoverable, subspace $D$ must intersect the manifold of real-world descriptors at multiple points. \cite{dusmanu2021privacy} proposes that half of the basis descriptors be randomly selected from a database of real-world descriptors $W$, and the other half be randomly generated via random sampling from a uniform distribution, \ie, setting $d_0=d$ and $d_i=a_i-d$ for $i=\{1,...,\frac{m}{2}\}$, where $a_i \sim \mathcal{U}\{W\}$ and $d_i \sim \mathcal{U}([-1,1])^n$ for $i=\{\frac{m}{2}{+}1,...,m\}$ . This way, $d$ and $\{a_1,...,a_\frac{m}{2}\}$ are contained in $D$.
\cite{dusmanu2021privacy} refers to this half-and-half approach as \textit{hybrid lifting}.

\noindent \textbf{Re-parameterization.} Evidently, the above representation of $D$ directly exposes the descriptors, hence \cite{dusmanu2021privacy} re-parameterizes $D$ to prevent information leakage. First, to avoid setting the  translation vector as the raw descriptor $d$, it randomly generates a new translation vector  $d_0=p_{\bot}^D(e_0)$, where $p_{\bot}^D(e_0)$ denotes the orthogonal projection of $e_0$ onto $D$, $e_0\sim \mathcal{U}([-1,1])^n$. Further, to prevent an attacker from using the direction of the basis descriptors to infer the raw descriptor $d$, a new set of basis descriptors $d_i=p_{\bot}^D(e_i)$ for $i=\{1,...,m\}$, where $e_i\sim \mathcal{U}([-1,1])^n$, are randomly generated. Note that the above two steps only re-parameterize $D$ without changing its intrinsic property.

\noindent \textbf{Matching.} With the lifted privacy-preserving representation, \cite{dusmanu2021privacy} further proposes the use of point-to-subspace and subspace-to-subspace distances for matching raw-to-lifted and lifted-to-lifted descriptors, respectively.

\section{Database and Clustering Inversion Attacks} 
\label{sec:inv_attacks}

In this section, we present two attacks against adversarial lifting \cite{dusmanu2021privacy}, namely the database attack and the clustering attack. In database attack,  we assume the attacker has access to the database of real-world descriptors $W$ from which the adversarial descriptors were selected, whereas in clustering attack, the attacker has no access to the database. Both attacks are based on the following key empirical assumption: a low-dimensional hybrid adversarial affine subspace $D$ likely only intersects the high-dimensional descriptor manifold at $\frac{m}{2}{+}1$ points corresponding to the original descriptor $d$ and the adversarial descriptors $\{a_1,...,a_{\frac{m}{2}}\}$ that were sampled from the database.

\begin{figure*}[th!]
  \centering
  \vspace{-0.4cm}
  \includegraphics[width=1.0\linewidth, trim = 0mm 65mm 0mm 0mm, clip]{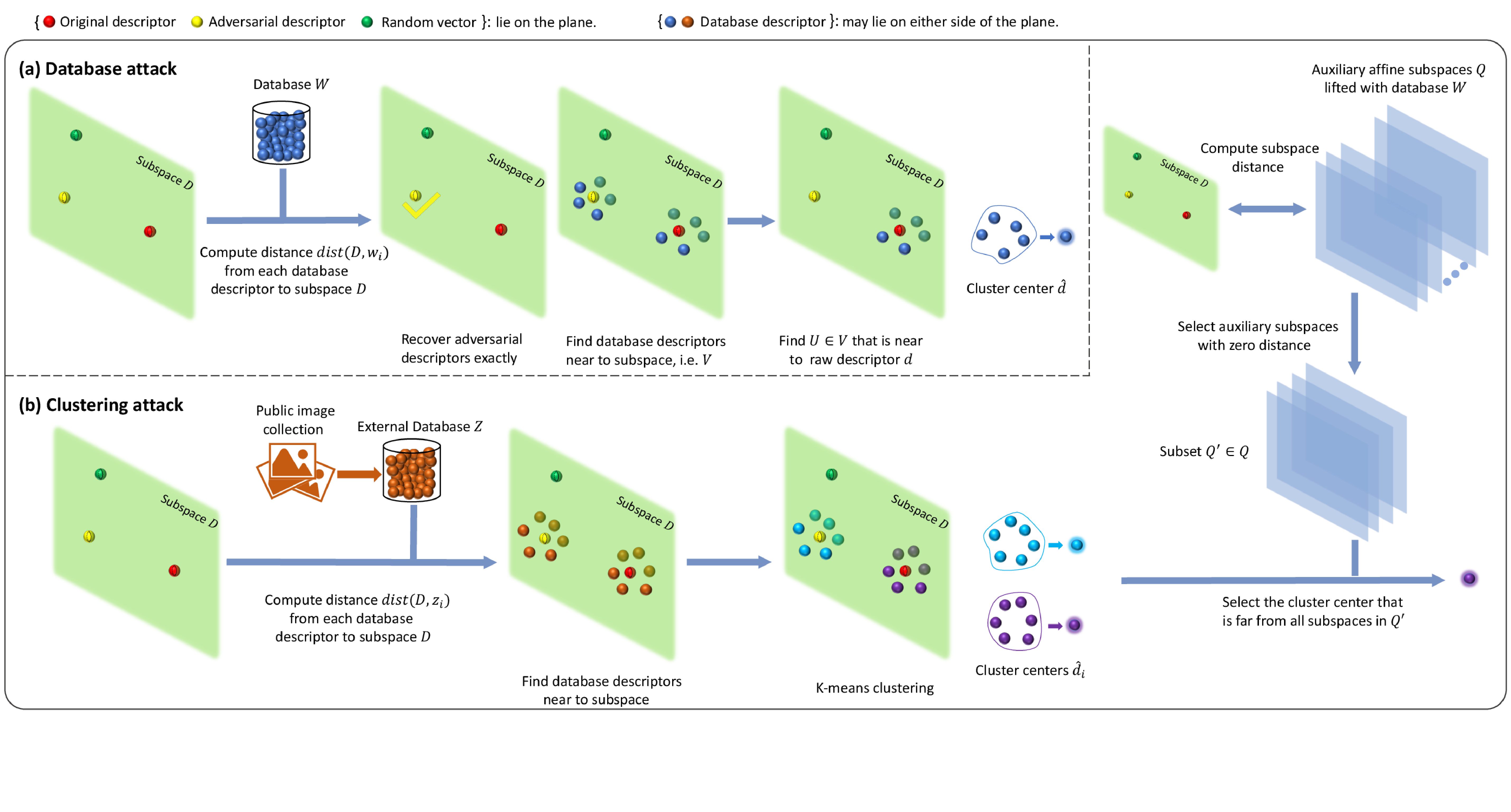}
  \centering
  \vspace{-0.2cm}
  \caption{\textbf{Illustration of (a) database attack and (b) clustering attack}.}  
  \vspace{-0.2cm}
  \label{fig:toyexample}
\end{figure*}

\subsection{Database Attack}
\label{sec:databaseattack}
With the above assumption, if we can identify the $\frac{m}{2}$ subspace-manifold intersections corresponding to the adversarial descriptors $\{a_1,...,a_{\frac{m}{2}}\}$, the recovery of descriptor $d$ from subspace $D$ is reduced to finding the one remaining intersection. This is exactly the outline of our attack, which we illustrate in \cref{fig:toyexample}(a) by a toy example with $n{=}3$ and $m{=}2$, \ie subspace being a plane in $\mathbb{R}^3$.

\noindent \textbf{Step 1: Compute distances to the database $W$.} We start by computing the distances $\mathtt{dist}(D,w_i) {=} ||w_i {-} p_{\bot}^D(w_i)||_2$ between subspace $D$ and each descriptor in the database of real-world descriptors $w_i {\in} W$, and then sort the descriptors in ascending order according to their respective distances. 

\noindent \textbf{Step 2: Recover adversarial descriptors exactly.} Recall that the adversarial descriptors are selected from $W$, thus $\mathtt{dist}(D,w_i){=}0$ holds exactly for $i{=}\{1,...,\frac{m}{2}\}$, which means 
the first $\frac{m}{2}$ descriptors from the sorted list immediately give our estimates $\{\hat{a}_1,...,\hat{a}_{\frac{m}{2}}\}$ for $\{{a}_1,...,{a}_{\frac{m}{2}}\}$.

\noindent \textbf{Step 3: Estimate the concealed descriptor.} Unlike the adversarial descriptors, the database does not contain the original descriptor $d$, but we may estimate it by its close neighbors in the database. To this end, the next $|V|$ descriptors in the sorted list, $V=\{v_1,...,v_{|V|}\}$, $|V|<<|W|$, are selected, containing the descriptors nearest to $D$ for which $\mathtt{dist}(D,v_i)>0$. More specifically, these descriptors are near to either the adversarial descriptors or the raw descriptor, and we aim to further select a subset $U \in V$ which is close to the raw descriptor $d$ but far from the adversarial ones. $U$ and $V$ are both illustrated in Fig.~\ref{fig:toyexample}(a). To select $U$, a score $s_i=\min_{j=1,...,\frac{m}{2}} ||\hat{a}_j - v_i||_2$ is computed for each $v_i$ in $V$. The descriptors with the highest scores, ${u_1,...,u_{|U|}}$, are used to estimate $d$ via weighted average and orthogonal projection:
\begin{equation}
    \hat{d} = p_{\bot}^D\Bigg(\frac{1}{\alpha} \sum_{i=1}^U  \frac{u_i}{\mathtt{dist}(D,u_i)} \Bigg),
\label{eq:clustercenter}
\end{equation}
where $\alpha = \sum_{i=1}^U \mathtt{dist}(D,u_i)^{-1}$. 

\noindent \textbf{Remark.} The intuition for why this attack works is that any descriptors from database $W$ that are near the subspace $D$ will likely cluster around either the original descriptor $d$ or one of the adversarial descriptors, as these are likely to be the only points where the subspace intersects the manifold of real-world descriptors. The effectiveness of this attack is empirically validated in \cref{sec:exp_inv}.

\subsection{Clustering Attack}
For this attack, we assume that the attacker does not have access to the database of real-world descriptors $W$ from which adversarial descriptors $a_{i=1,...,\frac{m}{2}}$ are sampled, but does have access to an additional set of adversarial affine subspaces $\mathcal{Q}$ that were lifted with the same database $W$. They could be obtained either from other descriptors in the same image or from a set of $c$ other images. The clustering attack is illustrated in \cref{fig:toyexample}(b).

\noindent \textbf{Step 1: Compute distance to public database.}  
Extract descriptors from a large set of public images and then cluster them to generate a public database of real-world descriptors $Z$ to serve as a proxy for the private database $W$. Then, compute the distances $\mathtt{dist}(D,z_i) = ||z_i - p_{\bot}^D(z_i)||_2$ between subspace $D$ and each descriptor $z_i \in Z$. 

\noindent \textbf{Step 2: Select nearest neighbors.}
Select the $|V|\ll|Z|$ descriptors nearest to subspace $D$, denoted $v_{i=1,...,V}$. Note, unlike in the database attack, we can't identify $a_{i=1,...,\frac{m}{2}}$ exactly, as, in general, $\mathtt{dist}(D,z_i) \neq 0$ for any $i$. 

\noindent \textbf{Step 3: Estimate candidate descriptors.} Cluster $V$ into $k=\frac{m}{2}+1$ clusters and assign each $v_i$ a cluster label $l_i \in \{1,...,\frac{m}{2}+1\}$.  Similar to \cref{eq:clustercenter}, the center of each cluster is computed separately as follows:
\begin{equation}
    \hat{d_i} = p_{\bot}^D\Bigg(\frac{1}{\alpha_i} \sum_{j=1}^V  \mathbbm{1}(l_j,i)\frac{v_i}{\mathtt{dist}(D,v_i)} \Bigg),
\end{equation}
where $\alpha_i = \sum_{j=1}^V \mathbbm{1}(l_j,i)\mathtt{dist}(D,j_i)^{-1}$ and $\mathbbm{1}(x,y)=1$ when $x=y$ and $0$ otherwise. 
Note, these $k$ descriptors represent (approximately) the intersections between subspace $D$ and the manifold of real-world descriptors. Thus, for our attack, we assume that each descriptor $\hat{d_i}$ is near to either the original descriptor $d$ or one of the adversarial descriptors $\{a_1,...,a_{\frac{m}{2}}\}$. Next, we leverage the auxiliary subspace $\mathcal{Q}$ to estimate which $\hat{d_i}$ is nearest to $d$. 

\noindent \textbf{Step 4: Estimate the concealed descriptor.} Recall that subspace $D$ and the subspaces in $\mathcal{Q}$  were lifted using the same database of private descriptors $W$. Thus, it's likely that a subset of subspaces $\mathcal{Q}' \in \mathcal{Q}$ were lifted using one or more of the same adversarial descriptors as $D$, \ie one of $a_i,i{=}\{1,...,\frac{m}{2}\}$. We can identify this subset $\mathcal{Q}'$ by noting that each subspace in $\mathcal{Q}'_j \in \mathcal{Q}'$ intersects with $D$, i.e., by selecting all $\mathcal{Q}_j \in \mathcal{Q}$ for which $dist(D,\mathcal{Q}_j)=0$. Assuming that $\mathcal{Q}$ is sufficiently large such that all $a_i$'s were used to lift at least one of the subspaces in $\mathcal{Q}'$, this indicates that the minimal point-to-subspace distance $\min_j \mathtt{dist}({a}_i,Q'_j) = 0$ for $i{=}\{1,...,\frac{m}{2}\}$. On the other hand, since $\mathcal{Q}'$ is selected without any knowledge or specific treatments on $d$, it is expected that  $\min_j \mathtt{dist}(d,Q'_j)\gg0$. In this discrepancy lies the crux of our attack --
while $\min_j \mathtt{dist}(\hat{a}_i,Q'_j)>0$ for our estimates of $a_i$, $\hat{a}_i$, we expect that $\min_j \mathtt{dist}(\hat{d},Q'_j) \gg \min_j \mathtt{dist}(\hat{a}_i,Q'_j)$. Hence, we compute the score $s_i$ for each $\hat{d}_i$ as  $s_i = \min_j \mathtt{dist}(\hat{d}_i,Q'_j)$ and the largest $s_i$ yields our estimate for $d$. We note that it is not impossible that $\mathcal{Q}'$ may contain a database descriptor that is close to $d$ too, but the probability of such a collision is low thanks to the high dimension of descriptors and empirically, our attack remains effective, as shown in \cref{sec:exp_inv}.

\vspace{0.2cm}
\noindent \textbf{Image reconstruction attack:} With the recovered raw descriptors by our database/clustering attack, one may perform an image inversion attack of the sort described in \cite{pittaluga2019revealing}.

\section{Our \ldpdesc}
\label{sec:im-ldp}

The success of our inversion attacks motivates the need for an image feature privatization method with rigorous privacy guarantee. We present the first solution towards this goal with local differential privacy. 

\subsection{Preliminary: Local Differential Privacy} 
Unlike original differential privacy~\cite{dwork2006differential}, the local differential privacy (LDP) setting allows users to sanitize their data locally before sending to a curator, so the curator needs not be trusted. Here, we describe necessary definitions of LDP and refer readers to \cite{xiong2020comprehensive} for detailed derivations.

\vspace{0.2cm}
\noindent \textbf{Definition 1 (Local Differential Privacy)}
\textit{A randomized mechanism $\mathcal{M}$ satisfies $\epsilon$-local differential privacy ($\epsilon$-LDP), where $\epsilon \geq 0$, if and only if for any inputs $x_1$ and $x_2$, 
\begin{equation}
    \forall y\in Range(\mathcal{M}): \frac{\Pr[\mathcal{M}(x_1)=y]}{\Pr[\mathcal{M}(x_2)=y]} \leq e^{\epsilon},
\label{eq:ldp}
\end{equation}}
\noindent where $Range(\mathcal{M})$ denotes the set of all possible outputs of $\mathcal{M}$. Note that $\mathcal{M}$ maps the input to a probability distribution rather than a single point. The $\epsilon$ controls the similarity in the output, and is termed as the \textit{privacy budget} -- a smaller $\epsilon$ indicates higher privacy protection, and vice versa. To illustrate this, we note that according to the definition of LDP, \cref{eq:ldp} holds too if we swap $x_1$ and $x_2$, \ie $\Pr[\mathcal{M}(x_2){=}y] \leq e^{\epsilon} \Pr[\mathcal{M}(x_1){=}y]$. When $\epsilon{=}0$, it follows that $\Pr[\mathcal{M}(x_2){=}y]{=}\Pr[\mathcal{M}(x_1){=}y]$. This means $x_1$ and $x_2$ have an identical distribution after $\mathcal{M}$ perturbation, and are indistinguishable from each other, hence yielding strongest privacy protection. Conversely, a larger $\epsilon$ loosens the constraint in \cref{eq:ldp} and reduces privacy protection.

\noindent \textbf{Selection of $\epsilon$.} While $\epsilon$ may be set as any value,  it is commonly set within $[0.01, 10]$, which was shown to ensure good privacy protection in practice ~\cite{hsu2014differential,yang2020local}.

\vspace{0.2cm}
\noindent \textbf{Definition 2 ($\omega$-Subset Mechanism)}
Denoting the data domain by $\mathcal{K}$, for any input $v \in \mathcal{K}$, randomly report a $\omega$-sized subset $\mathcal{Z}$ of $\mathcal{K}$, \ie $\mathcal{Z} {\subset} \mathcal{K}$ and $|\mathcal{Z}|{=}\omega$, with probability
\begin{equation}
\Pr(\mathcal{Z}|v) =
\begin{cases}
\frac{\omega e^{\epsilon}}{\omega e^{\epsilon} + |\mathcal{K}| - \omega} / {|\mathcal{K}| \choose \omega}, & \text{if } v \in \mathcal{Z}, \\
\frac{\omega}{\omega e^{\epsilon} + |\mathcal{K}| - \omega} / {|\mathcal{K}| \choose \omega}, & \text{if } v \notin \mathcal{Z}.
\end{cases}
\label{eq:wsm}
\end{equation}
The $\omega$-Subset Mechanism ($\omega$-SM) satisfies $\epsilon$-LDP \cite{wang2020comprehensive, yang2020local}. It is important to note that the data domain $\mathcal{K}$ is required to be a  finite space, \ie consisting of countably many elements. In what follows, we formulate our image feature perturbation as $\omega$-SM for privacy guarantee.

\subsection{Image Feature Matching with LDP}
\label{sec:im_ldp}

\noindent \textbf{Overview~~} Similarly to affine subspace lifting~\cite{dusmanu2021privacy}, we apply LDP to perturb feature descriptors before sending to the curator, and the curator applies feature matching with RANSAC-based geometric verification to enable downstream tasks despite the perturbation. We note the privacy-utility trade-off here -- a larger perturbation increases privacy protection but causes more significant challenges for correct matching. This trade-off is controlled in our framework by $\epsilon$, which corresponds to a guaranteed bound of privacy loss. Next, we present in detail our LDP protocol for image features by leveraging the $\omega$-Subset Mechanism.

\vspace{0.1cm}
\noindent \textbf{Naive Approach (LDP on the full descriptor space).} A straightforward approach is to apply $\omega$-SM directly on the descriptor space for obfuscation -- define $\mathcal{K}$ as the set of all possible descriptors and randomly report a subset of descriptors, which contain the raw descriptor with some probability. This is applicable to image descriptors as they have a finite domain size $|\mathcal{K}|$ as required by $\omega$-SM: $|\mathcal{K}|=2^{8\times128}$ for 128-dim uint8-based descriptors (\eg SIFT), and $|\mathcal{K}|=2^{32\times128}$ for 128-dim float32-based descriptors (\eg HardNet~\cite{mishchuk2017working}). However, as we shall demonstrate in  \cref{sec:exp}, naively setting the output space to the full descriptor domain does not lead to a desirable privacy-utility trade-off. This is caused by the domain size being too large; we will explain this shortly after introducing the following domain with a smaller size.  

\begin{figure*}
\centering
\includegraphics[width=\linewidth]{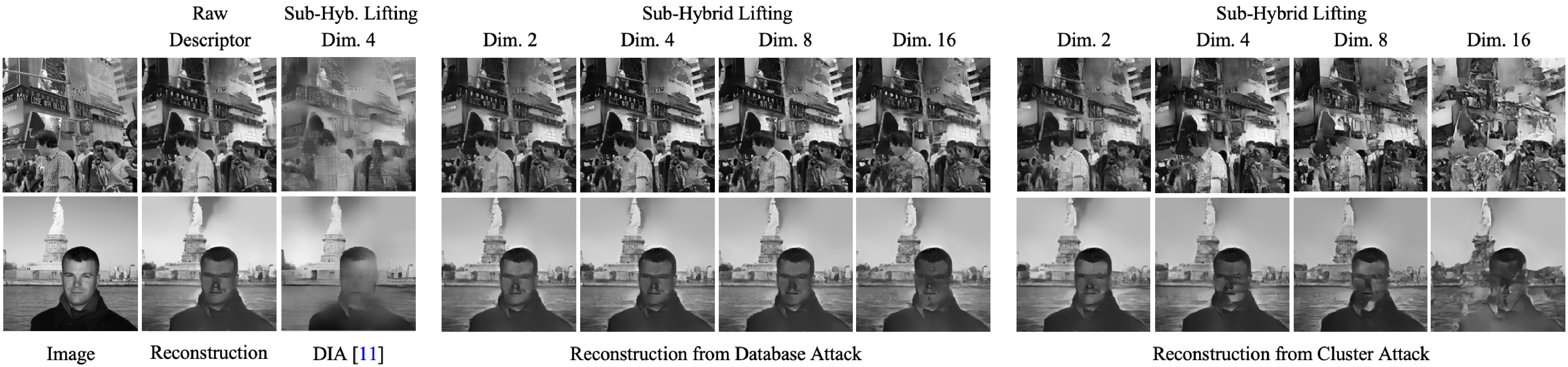}
\caption{\textbf{Inverting Adversarial Affine Subspace Embeddings.}}
\label{fig:attacks}
\end{figure*} 

\vspace{0.1cm}
\noindent \textbf{Our \ldpdesc (LDP on a dictionary of descriptors).} We instead define the data domain $\mathcal{K}$ as a finite dictionary of descriptors established from real-world image collections; this dictionary serves as the database shared with all users. More specifically, the database is created by extracting descriptors from a large public database of images and then performing k-means, as in \cite{dusmanu2021privacy}. Locally, each user enforces differential privacy by the following steps. \\
\noindent \textbf{Step 1:} Replace each descriptor $d$ that is to be sent to the curator with its nearest neighbor $d' \in \mathcal{K}$. \\
\noindent \textbf{Step 2:} Then, $d'$ is replaced with a set of descriptors $\mathcal{Z} \subset \mathcal{K}$ of size $m$. We perform random sampling~\cite{wang2020comprehensive} to generate the set $\mathcal{Z}$ that satisfies the probability distribution in \cref{eq:wsm}: first sample a Bernouilli scale variable $u$ with 
\begin{equation}
Pr(u=1) = \frac{m e^{\epsilon}}{m e^{\epsilon} + |\mathcal{K}| - m}; 
\label{eq:bernouilli}
\end{equation}
then randomly sample $m-u$ descriptors $\mathcal{Y}$ from $\mathcal{K}-\{d'\}$. \\
\noindent \textbf{Step 3:} if $u=1$, $\mathcal{Z}=\mathcal{Y} \cup \{d'\}$ else $\mathcal{Z}=\mathcal{Y}$.\\
\noindent This approach satisfies $\epsilon$-LDP (See supplementary for proof). Note that the curator will receive multiple descriptors per keypoint. 
The curator can then discover which, if any, of the matches for a given keypoint are correct by performing RANSAC-based geometric verification. Despite the perturbation on the descriptor, good empirical performance is still observed in downstream tasks, as shown in Sec.~\ref{sec:exp}.

\noindent \textbf{Why domain size matters?} Referring to \cref{eq:bernouilli}, it is clear that, with a fixed value of $\epsilon$,  an extremely large value of $|\mathcal{K}|$ renders $Pr(u=1)$ extremely small -- which severely limits the sending of raw descriptors to the server. This implies a very low proportion of inlier correspondences, which hinders utility. On the other hand, one may observe that increasing $\epsilon$ in tandem with $|\mathcal{K}|$ may prevent $Pr(u=1)$ from dropping, however, a larger $\epsilon$ quickly reduces the strength of privacy protection; recall that $\epsilon$ typically is confined within $[0.01,10]$ for practical usage~\cite{hsu2014differential,yang2020local}. As such, too large a domain size may cause poor privacy-utility trade-off. Similarly, increasing $m$, the number of descriptors sent to the server, prevents $Pr(u=1)$ from dropping at the cost of reducing the proportion of inliers, which inhibits utility too. These factors motivate our design choice to adopt a dictionary of descriptors-based approach. 

\noindent \textbf{Theorem 1 (\ldpdesc satisfies $\epsilon$-LDP).} Our \ldpdesc is not strictly a $\omega$-SM per se, because of the preceding nearest neighbor mapping -- we first map the raw input $d$ to its nearest neighbor $d'$ in the database $\mathcal{K}$, and then apply $\omega$-SM on top of $d'$. We prove in supplementary that \ldpdesc still satisfies $\epsilon$-LDP.

\vspace{0.2cm}
\noindent \textbf{Relation to affine subspace embeddings.~} While our method is similar to \cite{dusmanu2021privacy} in that users obfuscate descriptors by hiding them among a set of confounder descriptors randomly sampled from a database, there are two critical differences. Firstly, in our method, the set of descriptors $\mathcal{Z}$ sent by each user to the curator must be a subset of finite vocabulary $\mathcal{K}$; recall that the original descriptor $d$, if included, is also replaced by its nearest neighbor in $\mathcal{K}$. Hence, even if $\mathcal{K}$ is exactly known by a malicious curator, he cannot use $\mathcal{K}$ to perform a database attack of the sort described in \cref{sec:databaseattack}, and the same holds for the clustering attack when $\mathcal{K}$ is not accessible. Secondly, thanks to careful design of the obfuscation protocol, our method enables rigorous accounting of privacy leakage via local differential privacy, with a guaranteed bound of privacy leakage irrespective of the strength of attacks.

\section{Experimental Results}
\label{sec:exp}

\begin{figure*}
\centering
\includegraphics[width=\linewidth]{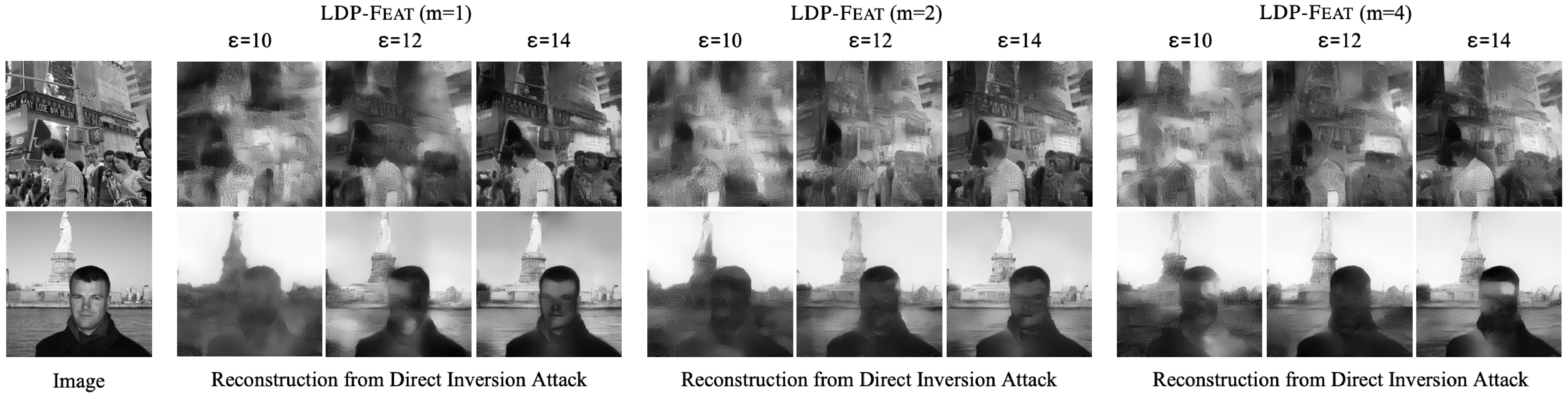}
\caption{\textbf{Direct Inversion Attack on \ldpdesc.} ($|\mathcal{K}|=256k$).}
\label{fig:attacks_ldp}
\end{figure*} 

In this section, we first evaluate the efficacy of our database and clustering attack, and then evaluate our \ldpdesc with respect to both utility and privacy.

\subsection{Inversion Attacks on Adversarial Affine Subspace Embeddings}
\label{sec:exp_inv}

\noindent \textbf{Evaluation setup.~} Our experimental setup is similar to \cite{dusmanu2021privacy}. For all evaluations, we employ sub-hybrid adversarial lifting of SIFT descriptors using an adversarial lifting database that was generated by clustering 10 million local features from 60,000 images from the Places365 dataset \cite{zhou2017places} into 256,000 clusters using spherical k-means \cite{bishop2006pattern}. As in \cite{dusmanu2021privacy}, the 256,000 descriptors in the adversarial database are split into 16 sub-databases. As described in \cref{sec:inv_attacks}, in \textit{database attack}  the adversary has access to this exact adversarial lifting database, whereas in \textit{clustering attack}, the adversary only has access to a public 
database of 128,00 descriptors, that was generated using the same process as above, but from a different set of 60,000 images. Additionally, as in \cite{dusmanu2021privacy}, we train a U-Net \cite{ronneberger2015u} style CNN for image reconstruction from descriptors on the MegaDepth dataset \cite{MDLi18} using the same architecture and loss as \cite{pittaluga2019revealing}. For all reconstructions, we report the following image reconstruction quality metrics: mean absolute error (MAE), structural similarity index measure (SSIM), and peak signal-to-noise ratio (PSNR).

\begin{table}[t]
\centering
\centerline{\resizebox{0.85\columnwidth}{!}{
\begin{tblr}{c|c|ccc}
  \toprule
  \textbf{Attack} & \textbf{Dim.} & \textbf{MAE} $\downarrow$ & \textbf{SSIM} $\uparrow$ & \textbf{PSNR} $\uparrow$ \\
  \midrule
  \SetCell[r=4]{c} Nearest \cite{dusmanu2021privacy} & 2 & .1690 & .3611 & 14.00 \\ 
      & 4 & .1913 & .3273 & 12.84 \\
      & 8 & .1985 & .2481 & 12.29  \\ 
      & 16 & .1873 & .2457 & 12.76  \\ 
  \midrule
  \SetCell[r=4]{c} DIA \cite{dusmanu2021privacy} & 2 & .1194 & .5005 & 16.35 \\ 
      & 4  & .1468 & .4190 & 14.79 \\
      & 8  & .1635 & .3676 & 14.01 \\ 
      & 16 & .1761 & .3496 & 13.43  \\ 
  \midrule 
  \SetCell[r=4]{c} Clustering (Ours) & 2 & .1087 & .5919 & 17.30 \\ 
      & 4 & .1142 & .5652 & 16.51 \\ 
      & 8 & .1254 & .5160 & 15.83  \\ 
      & 16 & .1424 & .4122 & 14.73  \\ 
  \midrule
  \SetCell[r=4]{c} Database (Ours) & 2 & .0947 & .6566 & 18.26 \\ 
      & 4 & .0950 & .6543 & 18.27  \\ 
      & 8 & .1000 & .6385 & 17.89 \\ 
      & 16 & .1063 & .5882 & 17.47  \\ 
  \midrule
  Raw  & n/a & .0913 & .6878 & 18.59 \\
  \bottomrule
  \end{tblr}
}}
\caption{\textbf{Inverting Adversarial Affine Subspace Embeddings.}}
\label{table:attacks}
\end{table}

\begin{table}[th!]
\centering
\centerline{\resizebox{\columnwidth}{!}{
\begin{tblr}{c|ccc|ccc}
\toprule
\SetCell[c=1]{c}{} & \textbf{DB Size} & \textbf{Privacy} & \textbf{\# Desc.} & \SetCell[c=3]{c}{\textbf{Day}} \\
& $|\mathcal{K}|$ & $\epsilon$ & $m$ & 0.25m, 2\degree & 0.5m, 5\degree  & 5.0m, 10\degree  \\
\midrule
\SetCell[r=5]{c} \textbf{\thead{Accuracy\\Upper\\Bound}} 
& $2^{1024}$ & $\infty$ & 1 & 84.1 & 91.7 & 96.4  \\
& 1024k      & $\infty$ & 1 & 79.7 & 89.9 & 94.9 \\
& 512k       & $\infty$ & 1 & 78.0 & 87.4 & 93.3  \\
& 256k       & $\infty$ & 1 & 76.8 & 86.3 & 91.6  \\
& 128k       & $\infty$ & 1 & 73.2 & 82.8 & 88.1  \\
\midrule
\SetCell[r=5]{c} \textbf{\thead{Impact of\\Database\\Size}}
& $2^{1024}$ & 10 & 2 & 0.00 & 0.00 & 0.00 \\
& 1024k      & 10 & 2 & 33.3 & 37.1 & 42.1 \\
& 512k       & 10 & 2 & 37.4 & 43.2 & 48.1 \\
& 256k       & 10 & 2 & 42.1 & 49.3 & 54.4 \\
& 128k       & 10 & 2 & 39.4 & 45.4 & 50.2 \\
\midrule
\SetCell[r=8]{c} \textbf{\thead{Privacy\\Guarantee}}
& 512k       & 16 & 4 & 76.1 & 85.0 & 90.4 \\
& 512k       & 14 & 4 & 73.9 & 84.5 & 90.2 \\
& 512k       & 12 & 4 & 69.4 & 77.9 & 84.6 \\
& 512k       & 10 & 4 & 42.1 & 49.6 & 55.0 \\
& 256k       & 16 & 2 & 75.4 & 85.3 & 90.2 \\
& 256k       & 14 & 2 & 75.1 & 84.7 & 89.4 \\
& 256k       & 12 & 2 & 69.5 & 78.4 & 84.1 \\
& 256k       & 10 & 2 & 42.1 & 49.3 & 54.4 \\
\midrule
\SetCell[r=5]{c} \textbf{\thead{Impact of\\Subset\\Size}} 
& 256k       & 10 & 1 & 34.6 & 39.7 & 44.7 \\
& 256k       & 10 & 2 & 42.1 & 49.3 & 54.4 \\
& 256k       & 10 & 4 & 42.1 & 49.0 & 55.1 \\
& 256k       & 10 & 8 & 39.1 & 46.4 & 51.8 \\
& 256k       & 10 &16 & 32.8 & 38.0 & 44.3 \\
\bottomrule
\end{tblr}
}}
\caption{\textbf{Aachen Day-Night Localization Challenge.}}
\label{table:loc}
\end{table}

\vspace{0.1cm}
\noindent \textbf{Image reconstruction attack.~} For this evaluation, a set of descriptor-keypoint pairs is extracted from a source image, transformed into a set of subspace-keypoint pairs using sub-hybrid adversarial lifting \cite{dusmanu2021privacy}, and then sent to the adversary. The goal of the adversary is to reconstruct the source image from the received subspace-keypoint pairs.
To achieve this, we first employ either the \textit{database attack} or the \textit{cluster attack} independently on each subspace-keypoint pair to recover an estimate of the original descriptor. The descriptors so obtained are then organized into a sparse feature map $F \in \mathbb{R}^{256 \times 256 \times 128}$ using the respective keypoint locations; pixels that do not have any associated descriptor are set to zero. Feature map $F$ is then provided as input to a pre-trained U-Net style reconstruction network to recover the original source image.

\vspace{0.1cm}
\noindent \textbf{Results.~} We evaluate the efficacy of the attacks on the 10 holiday images from Flickr selected by~\cite{dusmanu2021privacy}. For comparison, we also report the results of two other approaches from~\cite{dusmanu2021privacy}: \textit{Nearest} and \textit{Direct inversion attack (DIA)}. For \textit{Nearest}, each adversarial subspace is replaced with its nearest neighbor in the public database described above. For \textit{DIA}, the reconstruction network is trained to recover images directly from the affine subspace parameters.
As shown by the example qualitative results in \cref{fig:attacks}, our database attack is able to recover high-quality image content. The clustering attack lags slightly behind, but still reveals a significant amount of private information. We report quantitative reconstruction quality in  \cref{table:attacks}, for various lifting dimensions. 
As a reference for upper bound, we also present results from \textit{Raw}, where the original raw descriptor is input to the reconstruction network. As can be seen, both of our attacks are capable of recovering high-quality images from the estimated original descriptor, even for an adversarial dimension of 16.

\subsection{Applying \ldpdesc to Visual Localization}
As in \cite{dusmanu2021privacy}, we evaluate \ldpdesc on the task of visual localization on a pre-built map from the Aachen Day-Night long-term visual localization challenge \cite{VisualLocalization}. Following \cite{dusmanu2021privacy}, we privatize descriptors extracted from the query images, but not the reference images, to simulate an application aiming to protect the user privacy in an image-based localization service, such as Google Visual Positioning System \cite{Google2019Google} or Microsoft Azure Spatial Anchors \cite{Microsoft2019Announcing}. 

\vspace{0.15cm}
\noindent \textbf{Preliminaries.~} We generate the pre-built map by extracting SIFT descriptors from the reference images and triangulating the database model from the camera poses and intrinsics provided by \cite{VisualLocalization}. Next, we privatize the raw descriptors of all query images using \ldpdesc. For matching, we retrieve the top 20 reference images for each query image using ApGem \cite{revaud2019learning} and match the privatized descriptors to the reference descriptors. Finally, we obtain poses for the query images using the COLMAP \cite{schoenberger2016sfm} image registrator with fixed intrinsics. The poses are then submitted to the long-term visual localization benchmark. Below, we analyze the behavior of \ldpdesc with the evaluation metric being the percentages of localized query images for the day queries for different thresholds in the translation and rotation error, as shown in \cref{table:loc}. 

\begin{table}[t]
\centering
\resizebox{0.85\columnwidth}{!}{
\begin{tblr}{cc|ccc}
\toprule
\SetCell[r=2]{c} \textbf{Ablation} & \SetCell[r=2]{c} \textbf{Variable} & \SetCell[c=3]{c}{\textbf{Aachen Day}} \\
& & 0.25m, 2\degree & 0.5m, 5\degree  & 5.0m, 10\degree  \\
\midrule
\SetCell[r=2]{c} \textbf{\thead{Matching\\Algorithm}} 
& voc-match & 42.1 & 49.3 & 54.4 \\
& mutual-nn & 18.3 & 22.0 & 26.1 \\
\midrule
\SetCell[r=2]{c} \textbf{\thead{Public\\Database}}
& Aachen Ref. & 42.1 & 49.3 & 54.4 \\
& Places365 & 19.3 & 21.8 & 24.8 \\
\midrule
\SetCell[r=4]{c} \textbf{\thead{RANSAC\\Iterations}}
& 10M  & 42.1 & 49.3 & 54.4 \\
& 1M   & 40.2 & 45.6 & 51.3 \\
& 100k & 38.2 & 44.7 & 49.9 \\
& 10k  & 33.0 & 38.7 & 44.5 \\
\bottomrule
\end{tblr}
}
\caption{\textbf{Ablation Study.} ($|K|{=}256k$, $\epsilon{=}10$, $m{=}2$).}
\label{table:ablation}
\end{table}

\vspace{0.15cm}
\noindent \textbf{Localization accuracy upper bound.} For the first five rows in \cref{table:loc}, we set $m{=}1$ and $\epsilon{=}\infty$, meaning that \ldpdesc simply returns the nearest neighbor $d'$. Thus, these accuracies represent the upper bound performance for each database size. As expected, the localization accuracy decreases as the database size decreases, due to the quantization step in \ldpdesc that replaces a raw descriptor with its nearest neighbor in  $\mathcal{K}$. Note,  $|\mathcal{K}|{=}2^{1024}$ corresponds to our naive LDP method described in \cref{sec:im_ldp} that sets $\mathcal{K}$ as the set of all possible descriptors.

\vspace{0.15cm}
\noindent \textbf{Impact of database size.} Recall that \ldpdesc privatizes a raw descriptor $d$ by returning a random subset $\mathcal{Z} {\subset} \mathcal{K}$ of size $m$. Recall further that for a fixed $\epsilon$, the probability that $d' {\in} \mathcal{Z}$, where $d'$ denotes the nearest neighbor of $d$ in $\mathcal{K}$, is inversely related to the size of the  database $\mathcal{K}$. This creates an inherent privacy-utility tradeoff, as a larger database increases the resolution of the quantization step, but decreases the probability that $d' {\in} \mathcal{Z}$. To illustrate this, consider the following examples: (1) In the extreme case of $|\mathcal{K}|{=}m$, \ie always outputting the entire database $\mathcal{K}$ for any input, the privacy is fully preserved since none of the input is distinguishable in their output, but there is not any utility for feature matching; (2) In the case of $|\mathcal{K}|$ being extremely large, $\Pr(d'{\in}\mathcal{Z})=\Pr(u{=}1)$ is nearly zero, meaning the nearest neighbor of the raw descriptor may never be included in the output subset $\mathcal{Z}$, which, evidently, leads to poor utility. In \cref{table:loc}, we empirically investigate the impact of the database size on this tradeoff by fixing $\epsilon{=}10$ and $m{=}2$, and varying $|\mathcal{K}|$. Interestingly, we find that $|\mathcal{K}|{=}256k$ is the best operating point, which demonstrates the need to find the right quantization level.

\begin{table}[t]
\centering
\centerline{\resizebox{0.85\columnwidth}{!}{
\begin{tblr}{ccc|ccc}
  \toprule
 \textbf{DB Size}  & \textbf{Privacy} &  \textbf{\# Desc.} &\textbf{MAE} & \textbf{SSIM} & \textbf{PSNR} \\
  $|\mathcal{K}|$ & $\epsilon$ &  $m$ & ($\downarrow$) & ($\uparrow$) & ($\uparrow$) \\
  \midrule 
     256k & 16 & 1 & .1069 & .6248 & 17.35 \\
     256k & 14 & 1 & .1137 & .5923 & 16.90 \\
     256k & 12 & 1 & .1317 & .4668 & 15.77 \\ 
     256k & 10 & 1 & .1701 & .3651 & 13.72 \\ 
  \midrule 
     256k & 16 & 2 & .1173 & .5582 & 16.50 \\
     256k & 14 & 2 & .1208 & .5417 & 16.31 \\
     256k & 12 & 2 & .1366 & .4655 & 15.44 \\
     256k & 10 & 2 & .1660 & .3676 & 13.84 \\ 
  \midrule   
     256k & 16 & 4 & .1261 & .4976 & 15.94 \\ 
     256k & 14 & 4 & .1322 & .4868 & 15.47 \\
     256k & 12 & 4 & .1438 & .4511 & 14.93 \\
     256k & 10 & 4 & .1758 & .3666 & 13.48 \\
  \midrule  
     512k & 16 & 4 & .1226 & .4994 & 16.02 \\
     512k & 14 & 4 & .1254 & .4866 & 15.95 \\
     512k & 12 & 4 & .1511 & .4188 & 14.55 \\
     512k & 10 & 4 & .1685 & .3647 & 13.93 \\
  \bottomrule
  \end{tblr}
}}
\caption{\textbf{Direct Inversion Attack on \ldpdesc.}}
\label{table:attacks_ldp}
\end{table}

\vspace{0.15cm}
\noindent \textbf{Privacy guarantee.} The value of $\epsilon$ does not indicate the strength of privacy protection, but rather a \textit{bound} on the privacy leakage. As discussed above, the actual strength of the privacy protection of \ldpdesc depends not just on $\epsilon$, but also the database $\mathcal{K}$ and the subset size $m$. The unique advantage of \ldpdesc compared to existing visual privacy methods (\eg~\cite{dusmanu2021privacy}) is that \ldpdesc provides a privacy guarantee. We analyze in \cref{table:loc} the impact of varying $\epsilon$ on localization accuracy for two different operating points: (i) $(|\mathcal{K}|=512k, m=4)$ and  (ii) $(|\mathcal{K}|=256k, m=2)$. First, observe the decreasing accuracy with decreasing $\epsilon$. Interestingly, we further find that the two operating points achieve almost identical accuracy for the same $\epsilon$ value. A likely explanation for this is that the probability that $p(d' \in Z)$ is equal for both operating points, so the improved resolution of the operating point with larger $|\mathcal{K}|$ is negated by having to assign each keypoint $m=4$ instead of $m=2$ descriptors, in order to preserve the same $\epsilon$ value. Note, for a fixed value of $\epsilon$, we can search the parameters of $\mathcal{K}$ and $m$ for best utility while being assured that the privacy leakage is always bounded.

\vspace{0.15cm}
\noindent \textbf{Impact of Subset Size.} In the last five rows of \cref{table:loc}, we investigate the impact of varying the subset size of \ldpdesc. Empirically, we find that for a fixed database size $|K|=256k$ and $\epsilon=10$, the subset size $m=2$ and $m=4$ achieve the best localization accuracy. Again, this search is guarded by the privacy guarantee of \ldpdesc.

\vspace{0.15cm}
\noindent \textbf{Ablation Study.} We perform additional ablation experiments in \cref{table:ablation}, where we fix $|K|{=}256k$, $\epsilon{=}10$ and $m=2$. In the first two rows, we examine the impact of the matching criteria on localization accuracy and find that vocabulary-based matching \cite{nister2006} improves localization performance dramatically compared to mutual nearest neighbor. Next, we vary the source of the public database and find that extracting the descriptors from the Aachen reference images results in much better performance (more discussions below.). Finally, we report localization accuracy when varying the number of RANSAC iterations. As expected, more RANSAC iterations leads to better performance. 

\vspace{0.15cm}
\noindent \textbf{Impact of database content.} Following the above discussion, we further note that the database content in $\mathcal{K}$ also has an impact on the privacy-utility trade-off through the quantization step. Specifically, if the database descriptors are all significantly distinct from the input descriptor, it leads to large quantization errors as a result, indicating none of the database descriptors in the output subset $\mathcal{Z}$ can well represent the input descriptor. This certainly yields strong privacy but leads to poor utility. In \cref{table:ablation}, we have compared the performance of using Aachen reference images to build the database versus using the external Places365 dataset. The former achieves significantly superior performance due to its higher resemblance to the query images. Another example scenario where the database content plays a role lies in the Aachen night-time localization challenge. As we show in the supplementary, given night-time query images and the database $\mathcal{K}$ built solely from day-time images, the localization accuracy is degraded despite the strong privacy.

\subsection{Direct Inversion Attack on \ldpdesc.}
We start by noting that our database and clustering attack are not applicable to \ldpdesc since all descriptors sent are samples from the database. Instead, we evaluate the performance from a direct inversion attack. To achieve this, we stack all $m$ descriptors sent by the user and create the sparse feature map $F \in \mathbb{R}^{256 \times 256 \times (128 \times m)}$, similarly to what is described in \cref{sec:exp_inv}. $F$ is then fed as input to a U-Net style neural network trained to reconstruct the source image from such a feature map. As shown qualitatively in \cref{fig:attacks_ldp} with different combinations of $\epsilon$ and $m$, the attack generally fails to recover high-quality image content that reveals privacy. We further report quantitative results in \cref{table:attacks_ldp}. One observes that the image reconstruction quality is, in general, lower than that for the affine embeddings shown in \cref{table:attacks}, indicating the privacy protection of \ldpdesc.

\section{Conclusion}
In this paper, we propose two novel attacks to reveal the privacy leakage underneath the adversarial affine subspace embeddings~\cite{dusmanu2021privacy}. Following this, we further propose \ldpdesc, a new and more rigorous privacy-preserving protocol that formulates image feature matching under the umbrella of local differential privacy. This makes our method the first of its kind that enjoys the theoretical privacy guarantees offered by differential privacy, all the while achieving strong utility in downstream tasks, such as visual localization. We envision that our work will inspire more research efforts on specialized LDP protocols for image matching and other vision tasks.

{\small
\bibliographystyle{ieee_fullname}
\bibliography{main.bib}
}


\newpage
\clearpage

\setcounter{section}{0}
\setcounter{figure}{0}
\setcounter{table}{0}
\setcounter{equation}{0}
\setcounter{footnote}{0}
\renewcommand{\thepage}{S\arabic{page}}
\renewcommand{\thesection}{S\arabic{section}}
\renewcommand{\thefigure}{S\arabic{figure}}
\renewcommand{\thetable}{S\arabic{table}}
\renewcommand{\theequation}{S\arabic{equation}}

\begin{strip}
\begin{center}
\textbf{\Large Supplementary Material\\\ldpdesc: Image Features with Local Differential Privacy} \\
\vspace{2.5em}
{\large \quad Francesco Pittaluga \quad \quad \quad \quad \quad \quad \quad
Bingbing Zhuang}\\
{\tt\small francescopittaluga@nec-labs.com} \quad \quad \quad {\tt\small bzhuang@nec-labs.com} \\
{\large \quad NEC Labs America}\\
\end{center}
\end{strip}

\vspace{3em}
The supplementary material contains (1) a proof that \ldpdesc satisfies $\epsilon$-LDP, (2) additional experimental results on Aachen night-time localization and Structure-from-Motion (SfM), and (3) an analysis of the paper's assumptions.

\section{Local Differential Privacy of \ldpdesc}
Here, we prove that \ldpdesc satisfies $\epsilon$-LDP. 
For clarity, we first prove that $\omega$-SM satisfy $\omega$-LDP and the proof for \ldpdesc follows very similarly.

\subsection{Theorem 1 ($\omega$-Subset satisfies $\epsilon$-LDP)} 
For any inputs $v_1$ and $v_2$, and their output $\mathcal{Z}_1$ and $\mathcal{Z}_2$ returned by $\omega$-SM, there are four possible scenarios $\{v_1{\in}\mathcal{Z}_1,v_2{\in}\mathcal{Z}_2\}$, $\{v_1{\notin}\mathcal{Z}_1,v_2{\in}\mathcal{Z}_2\}$, $\{v_1{\in}\mathcal{Z}_1,v_2{\notin}\mathcal{Z}_2\}$, $\{v_1{\notin}\mathcal{Z}_1,v_2{\notin}\mathcal{Z}_2\}$, each with different probability distributions. Below, we show that the probability inequality required by $\epsilon$-LDP, \ie Eq.~(3) of the main paper, is satisfied for all the four scenarios. \\
1) $\{v_1{\in}\mathcal{Z}_1,v_2{\in}\mathcal{Z}_2\}$. In this case,  
\begin{equation}\begin{split}
\Pr(\mathcal{Z}_1|v_1) = \frac{\omega e^{\epsilon}}{\omega e^{\epsilon} + |\mathcal{K}| - \omega} / {|\mathcal{K}| \choose \omega}, \\
\Pr(\mathcal{Z}_2|v_2) = \frac{\omega e^{\epsilon}}{\omega e^{\epsilon} + |\mathcal{K}| - \omega} / {|\mathcal{K}| \choose \omega},
\end{split}
\end{equation} 
meaning that $\Pr(\mathcal{Z}_1|v_1){=}\Pr(\mathcal{Z}_2|v_2)$, hence $\Pr(\mathcal{Z}_1|v_1) \leq e^\epsilon \Pr(\mathcal{Z}_2|v_2)$ holds.\\ 
2) $\{v_1{\notin}\mathcal{Z}_1,v_2{\in}\mathcal{Z}_2\}$. In this case, 
\begin{equation}\begin{split}
\Pr(\mathcal{Z}_1|v_1) = \frac{\omega }{\omega e^{\epsilon} + |\mathcal{K}| - \omega} / {|\mathcal{K}| \choose \omega}, \\ \Pr(\mathcal{Z}_2|v_2) = \frac{\omega e^{\epsilon}}{\omega e^{\epsilon} + |\mathcal{K}| - \omega} / {|\mathcal{K}| \choose \omega},
\end{split}\end{equation} 
meaning that $\Pr(\mathcal{Z}_1|v_1)=e^{-\epsilon}\Pr(\mathcal{Z}_2|v_2)$, hence $\Pr(\mathcal{Z}_1|v_1) \leq e^\epsilon \Pr(\mathcal{Z}_2|v_2)$ holds since $\epsilon>0$. \\
3) $\{v_1{\in}\mathcal{Z}_1,v_2{\notin}\mathcal{Z}_2\}$.  In this case,
\begin{equation}\begin{split}
\Pr(\mathcal{Z}_1|v_1) = \frac{\omega e^{\epsilon}}{\omega e^{\epsilon} + |\mathcal{K}| - \omega} / {|\mathcal{K}| \choose \omega},\\
\Pr(\mathcal{Z}_2|v_2) = \frac{\omega }{\omega e^{\epsilon} + |\mathcal{K}| - \omega} / {|\mathcal{K}| \choose \omega},
\end{split}\end{equation} 
meaning that $\Pr(\mathcal{Z}_1|v_1)=e^{\epsilon}\Pr(\mathcal{Z}_2|v_2)$, hence $\Pr(\mathcal{Z}_1|v_1) \leq e^\epsilon \Pr(\mathcal{Z}_2|v_2)$ holds.
\\
4) $\{v_1{\notin}\mathcal{Z}_1,v_2{\notin}\mathcal{Z}_2\}$. In this cases,
\begin{equation}\begin{split}\Pr(\mathcal{Z}_1|v_1) = \frac{\omega}{\omega e^{\epsilon} + |\mathcal{K}| - \omega} / {|\mathcal{K}| \choose \omega}, \\ \Pr(\mathcal{Z}_2|v_2) = \frac{\omega}{\omega e^{\epsilon} + |\mathcal{K}| - \omega} / {|\mathcal{K}| \choose \omega}, 
\end{split}\end{equation} 
meaning that $\Pr(\mathcal{Z}_1|v_1){=}\Pr(\mathcal{Z}_2|v_2)$, hence $\Pr(\mathcal{Z}_1|v_1) \leq e^\epsilon \Pr(\mathcal{Z}_2|v_2)$ holds.\\ 
This concludes our proof.

\begin{table*}[t]
\centering
\vspace{-0.4cm}
\begin{tblr}{c|ccc|ccc|ccc}
\toprule[1.5pt]
& \textbf{DB Size} & \textbf{Privacy} & \textbf{\# Desc.} 
& & \textbf{Day} & & & \textbf{Night} & & \\
\SetCell[r=1]{c} & $|\mathcal{K}|$ & $\epsilon$ & $m$ & 0.25m, 2\degree & 0.5m, 5\degree  & 5.0m, 10\degree & 0.25m, 2\degree & 0.5m, 5\degree  & 5.0m, 10\degree  \\
\midrule
\SetCell[r=5]{c} \textbf{\thead{Accuracy\\Upper\\Bound}} 
& 128k       & $\infty$ & 1 & 73.2 & 82.8 & 88.1 & 24.6 & 28.8 & 33.5 \\
& 256k       & $\infty$ & 1 & 76.8 & 86.3 & 91.6 & 28.8 & 34.6 & 42.4 \\
& 512k       & $\infty$ & 1 & 78.0 & 87.4 & 93.3 & 33.5 & 40.8 & 51.3 \\
& 1024k      & $\infty$ & 1 & 79.7 & 89.9 & 94.9 & 36.1 & 42.4 & 51.8 \\
& $2^{1024}$ & $\infty$ & 1 & \textbf{84.1} & \textbf{91.7} & \textbf{96.4} & \textbf{50.3} & \textbf{61.8} & \textbf{73.8} \\
\midrule
\SetCell[r=5]{c} \textbf{\thead{Impact of\\Database\\Size}}
& 128k       & 10 & 2 & 39.4 & 45.4 & 50.2 & 1.60 & 2.10 & 3.70 \\
& 256k       & 10 & 2 & \textbf{42.1} & \textbf{49.3} & \textbf{54.4} & \textbf{3.70} & \textbf{5.20} & \textbf{6.30} \\
& 512k       & 10 & 2 & 37.4 & 43.2 & 48.1 & 2.10 & 3.70 & 4.70 \\
& 1024k      & 10 & 2 & 33.3 & 37.1 & 42.1 & 2.10 & 3.70 & 3.70 \\
& $2^{1024}$ & 10 & 2 & 0.00 & 0.00 & 0.00 & 0.00 & 0.00 & 0.00 \\
\midrule
\SetCell[r=8]{c} \textbf{\thead{Privacy\\Guarantee}}
& 256k       & 10 & 2 & 42.1 & 49.3 & 54.4 & 3.70 & 5.20 & 6.30 \\
& 256k       & 12 & 2 & 69.5 & 78.4 & 84.1 & 16.8 & 19.4 & 23.6 \\
& 256k       & 14 & 2 & 75.1 & 84.7 & 89.4 & \textbf{23.0} & \textbf{27.7} & \textbf{33.0} \\
& 256k       & 16 & 2 & \textbf{75.4} & \textbf{85.3} & \textbf{90.2} & \textbf{23.0} & 27.2 & 31.9 \\
\midrule
& 512k       & 10 & 4 & 42.1 & 49.6 & 55.0 & 5.20 & 7.90 & 9.90 \\
& 512k       & 12 & 4 & 69.4 & 77.9 & 84.6 & 19.4 & 22.5 & 26.2 \\
& 512k       & 14 & 4 & 73.9 & 84.5 & 90.2 & 23.6 & 29.8 & 34.6 \\
& 512k       & 16 & 4 & \textbf{76.1} & \textbf{85.0} & \textbf{90.4} & \textbf{24.6} & \textbf{29.8} & \textbf{34.0} \\
\midrule
\SetCell[r=5]{c} \textbf{\thead{Impact of\\Subset\\Size}}
& 256k       & 10 & 1 & 34.6 & 39.7 & 44.7 & 2.60 & 3.70 & 4.70 \\
& 256k       & 10 & 2 & \textbf{42.1} & \textbf{49.3} & 54.4 & \textbf{3.70} & \textbf{5.20} & \textbf{6.30} \\
& 256k       & 10 & 4 & \textbf{42.1} & 49.0 & \textbf{55.1} & \textbf{3.70} & 4.70 & 5.20 \\
& 256k       & 10 & 8 & 39.1 & 46.4 & 51.8 & 4.70 & 5.80 & 6.30 \\
& 256k       & 10 &16 & 32.8 & 38.0 & 44.3 & 2.60 & 2.60 & 3.70 \\
\bottomrule[1.5pt]
\end{tblr}
\caption{\textbf{Aachen Day-Night Localization Challenge.}}
\label{table:loc_night}
\end{table*}

\subsection{Theorem 2 (\ldpdesc satisfies $\epsilon$-LDP)} 
For any input descriptor $d$, the output set $\mathcal{Z}$ are obtained by:
first map $d$ to an element (let us denote it as $\bar{d}$) in the database $\mathcal{K}$, and then $\bar{d}$ is mapped to the random set $\mathcal{Z}$. Hence,

\begin{equation} \begin{split} \Pr(\mathcal{Z}|d) & = \sum_{\bar{d} \in \mathcal{K}}\Pr(\mathcal{Z}, \bar{d}|d)\\
& = \sum_{\bar{d} \in \mathcal{K}}\Pr(\mathcal{Z}|\bar{d})\Pr(\bar{d}|d).
\end{split}
\label{eq:zgivend}
\end{equation}
Since the mapping from $d$ to $\bar{d}$ is deterministic -- it is mapped to the nearest neighbor $d'$ in the database, we have 
\begin{equation}
\Pr(\bar{d}|d) =
\begin{cases}
1, & \text{if } \bar{d}=d, \\
0, & \text{if } \bar{d}\neq d.
\end{cases}
\label{eq:bardgived}
\end{equation}
Plugging \cref{eq:bardgived} into \cref{eq:zgivend} yields
\begin{equation} 
\Pr(\mathcal{Z}|d)= \Pr(\mathcal{Z}|d') 
\label{eq:zgivendfinal}
\end{equation}
For any input descriptor $d_1$ and $d_2$, their nearest neighbor $d_1'$ and $d_2'$, and their output $\mathcal{Z}_1$ and $\mathcal{Z}_2$, there are four possible scenarios 
$\{d_1'{\in}\mathcal{Z}_1, d_2'{\in}\mathcal{Z}_2\}$, $\{d_1'{\notin}\mathcal{Z}_1, d_2'{\in}\mathcal{Z}_2\}$, $\{d_1'{\in}\mathcal{Z}_1, d_2'{\notin}\mathcal{Z}_2\}$, 
$\{d_1'{\notin}\mathcal{Z}_1, d_2'{\notin}\mathcal{Z}_2\}$, each with different probability distributions. Since $\mathcal{Z}_1$ and $\mathcal{Z}_2$ are sampled using the $\omega$-SM, we have shown above that $Pr(\mathcal{Z}_1|d_1') \leq e^\epsilon Pr(\mathcal{Z}_2|d_2')$ holds for all the four scenarios, and given \cref{eq:zgivendfinal}, we have $Pr(\mathcal{Z}_1|d_1) \leq e^\epsilon Pr(\mathcal{Z}_2|d_2)$. This means that \ldpdesc satisfies $\epsilon$-LDP.

\begin{figure*}[t]
  \centering
  \begin{minipage}[c]{0.65\textwidth}
  \centerline{\resizebox{\textwidth}{!}{
    \begin{tblr}{c|ccc|cccc}
    \toprule[1.5pt]
    \SetCell[r=2]{c} \textbf{Scene} & \textbf{Dict. Size} & \textbf{Privacy} & \textbf{\# Desc.} & \textbf{Reg.}   & \textbf{Sparse} & \textbf{Track}  & \textbf{Reproj.} \\
    & $|\mathcal{K}|$ & $\epsilon$ & $m$ & \textbf{Images} & \textbf{Points} & \textbf{Length} & \textbf{Error} \\
    \midrule
    \SetCell[r=7]{c} \textcolor{black}{\textbf{\thead{South\\Building}}}
    &  $2^{1024}$ & $\infty$ & 1 & 128 & 110,714& 5.66 & 1.29 \\
    &  512k       & $\infty$ & 1 & 128 & 62,194 & 4.85 & 1.12 \\
    &  512k       & 10       & 4 & 88  & 8,668  & 3.56 & 0.85 \\
    &  512k       & 10       & 8 & 75  & 10,554 & 3.66 & 1.05 \\
    &  512k       & 10       & 16 & 123 & 25,451 & 3.62 & 0.89 \\
    &  512k       & 10       & 32 & 123 & 26,760 & 3.52 & 0.94 \\
    &  512k       & 10       & 64 & 124 & 21,596 & 3.31 & 1.28 \\
    \midrule
    \SetCell[r=7]{c} \textcolor{black}{\textbf{\thead{Fountain}}}
    & $2^{1024}$ & $\infty$ & 1 & 11 & 15,332& 4.42 & 2.82 \\
    & 512k       & $\infty$ & 1 & 11 & 8,612 & 3.80 & 2.39 \\
    & 512k       & 10       & 4 & 11 & 983  & 2.86 & 1.26 \\
    & 512k       & 10       & 8 & 11 & 1,827 & 2.98 & 1.35\\
    & 512k       & 10       & 16 & 11 & 2,598 & 3.03 & 1.50 \\
    & 512k       & 10       & 32 & 11 & 3,078 & 3.02 & 1.52 \\
    & 512k       & 10       & 64 & 11 & 3,242 & 2.89 & 1.41 \\
    \bottomrule[1.5pt]
    \end{tblr}
    }}
  \end{minipage}
  \begin{minipage}[c]{0.33\textwidth}
    \includegraphics[width=\textwidth]{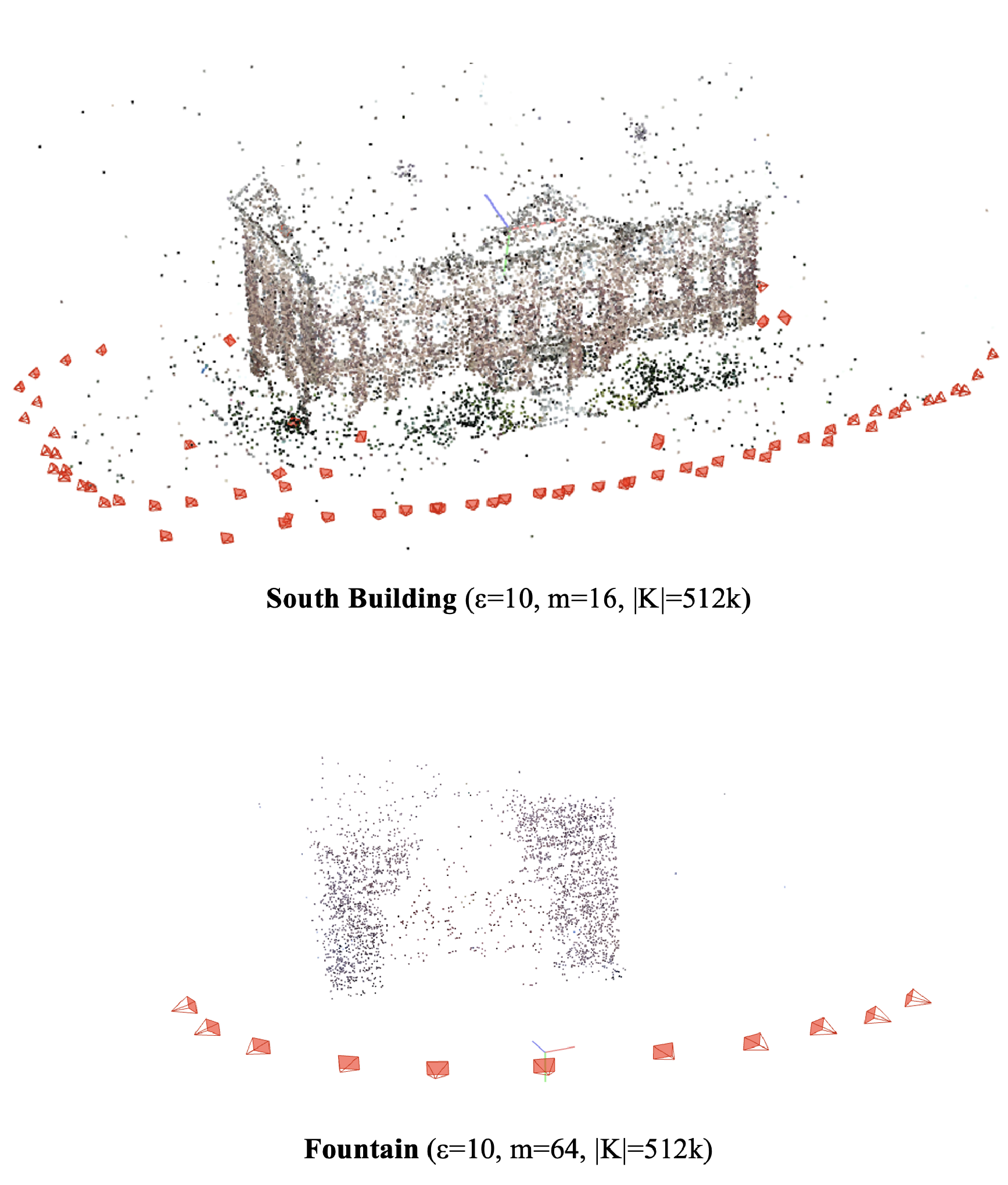}
  \end{minipage}
  \captionsetup{justification=centering}
  \caption{\textbf{Local Feature Evaluation Benchmark.} Structure-from-Motion results using \ldpdesc with different configurations. }
  \label{fig:sfm}
\end{figure*}

\section{Additional Results}
\subsection{Aachen Night Localization}

Similarly to the Tab.2 of the main paper, we report in \cref{table:loc_night} the localization accuracy for night-time queries in the Aachen Day-Night localization challenge. Overall, we observe a degradation of performance compared to the day-time queries. This is mainly because our database $\mathcal{K}$ was built from the Aachen reference images which contain day-time images only. As aforementioned, this causes a large quantization error $\Delta d$ in \ldpdesc, which certainly enhances privacy protection but compromises the utility. We leave the pursuit of a better privacy-utility trade-off for night-time localization as a future work.

\subsection{Structure-from-Motion}
We further demonstrate the utility of \ldpdesc on Structure-from-Motion, as shown in \cref{fig:sfm}. We adopt COLMAP~\cite{schoenberger2016sfm} for SfM by customizing its feature extraction and matching using \ldpdesc. As an indicator for SfM performance, we report the number of registered images, the number of reconstructed sparse 3D points, the average keypoint track length, and the average reprojection error.

We report results on the ``South Building" and ``Fountain" scene from the 3D reconstruction benchmark~\cite{schoenberger2017comparative}. We first report the results for ($|\mathcal{K}|=2^{1024}$, $\epsilon=\infty$, $m=1$). This corresponds to the oracle setting where only the raw descriptor is sent without any privacy protection, and which serves a performance upper bound. We then use a dictionary with $512k$ descriptors, \ie ($|\mathcal{K}|=512k$, $\epsilon=\infty$, $m=1$) where the quantization step, \ie mapping the raw descriptor $d$ to its nearest neighbor $d'$, introduces a degree of privacy protection and thus degrades the performance accordingly. Next, we fix $|\mathcal{K}|=512k$ and $\epsilon=10$, while increasing $m$ from 4 to 64. The performance varies, and we observe that $m{=}32$ yields the best performance. Overall, one observes that good SfM results are obtained from \ldpdesc under different settings; in particular, most of the cameras are successfully registered, despite the reconstructed points being sparser. We demonstrate the qualitative reconstruction results in \cref{fig:sfm}.

\section{Analysis of Assumptions}

\begin{table}[t]
\begin{center}
\begin{tblr}{c|cccc}
\toprule[1.5pt]
\textbf{Dim} & \SetCell[c=4]{c} \textbf{Success Rate (\%)}\\ 
$m$ & N=50 & N=20 & N=10 & N=5 \\ 
\midrule
4   & 93.89  & 95.60  & 96.61  & 97.46 \\
16  & 87.27  & 90.54  & 92.73  & 94.38 \\
\bottomrule[1.5pt]
\end{tblr}
\end{center}
\caption{\textbf{Intersecting Adversarial Subspaces.}}
\label{table:inters}
\end{table}

\subsection{Intersecting Adversarial Subspaces}

As discussed in the paper, our proposed Database and Clustering attacks are based on the following key empirical assumption: a low-dimensional hybrid adversarial affine subspace $D$ likely only intersects the high-dimensional descriptor manifold at $\frac{m}{2}{+}1$ points corresponding to the original descriptor $d$ and the adversarial descriptors $\{a_1,...,a_{\frac{m}{2}}\}$ that were sampled from the database. Here, we generate subspaces for 100K descriptors and report how often our assumption holds, i.e., for each subspace, we select the top N database descriptors closest to the subspace, and match them to the $\frac{m}{2}{+}1$ descriptors forming the subspace. Using the standard ratio test (${>}0.8$) we report the percentage of successful matches in \cref{table:inters}. The high success rates empirically validate our assumption regarding the rareness of intersections beyond the $\frac{m}{2}{+}1$ forming descriptors. We also note that our assumption is implied by the success of feature matching in \cite{dusmanu2021privacy}.

\begin{table}[t]
\begin{center}
\begin{tblr}{cccc}
\toprule[1.5pt]
Dim=2 & Dim=4 & Dim=8 & Dim=16 \\
\midrule
97.42\%  & 94.30\%  & 85.46\%  & 73.03\% \\
\bottomrule[1.5pt]
\end{tblr}
 \end{center}
 \caption{\textbf{Clustering Attack Collisions.}}
\label{table:collisions}
\end{table}

\subsection{Clustering Attack Collisions} 
For our clustering attack, we assume that the attacker does not have access to the database of real-world descriptors $W$ from which adversarial descriptors $a_{i=1,...,\frac{m}{2}}$ for subspace $D$ are sampled, but does have access to an additional set of adversarial affine subspaces $\mathcal{Q}$ that were lifted with the same database $W$. We can identify the subset of subspaces $\mathcal{Q}' \in \mathcal{Q}$ that were lifted using one or more of the same adversarial descriptors as $D$, by noting that each subspace in $\mathcal{Q}'_j \in \mathcal{Q}'$ intersects with $D$. Assuming that $\mathcal{Q}$ is sufficiently large such that all $a_i$'s were used to lift at least one of the subspaces in $\mathcal{Q}'$, this indicates that the minimal point-to-subspace distance $\min_j \mathtt{dist}({a}_i,Q'_j) = 0$ for $i{=}\{1,...,\frac{m}{2}\}$. On the other hand, since $\mathcal{Q}'$ is selected without any knowledge or specific treatments on $d$, it is expected that  $\min_j \mathtt{dist}(d,Q'_j)\gg0$. In this discrepancy lies the crux of our attack --
while $\min_j \mathtt{dist}(\hat{a}_i,Q'_j)>0$ for our estimates of $a_i$, $\hat{a}_i$, we expect that $\min_j \mathtt{dist}(\hat{d},Q'_j) \gg \min_j \mathtt{dist}(\hat{a}_i,Q'_j)$. Hence, we compute the score $s_i$ for each $\hat{d}_i$ as  $s_i = \min_j \mathtt{dist}(\hat{d}_i,Q'_j)$ and the largest $s_i$ yields our estimate for $d$. We note that it is not impossible that $\mathcal{Q}'$ may contain a database descriptor that is close to $d$ too, but the probability of such a collision is low thanks to the high dimension of descriptors. In \cref{table:collisions}, we validate this assumption by lifting all the descriptors of our 10 test images to adversarial subspaces and reporting the percentage of them that have no collisions in our attack.

\begin{figure}[t]
\centering
\includegraphics[width=\linewidth]{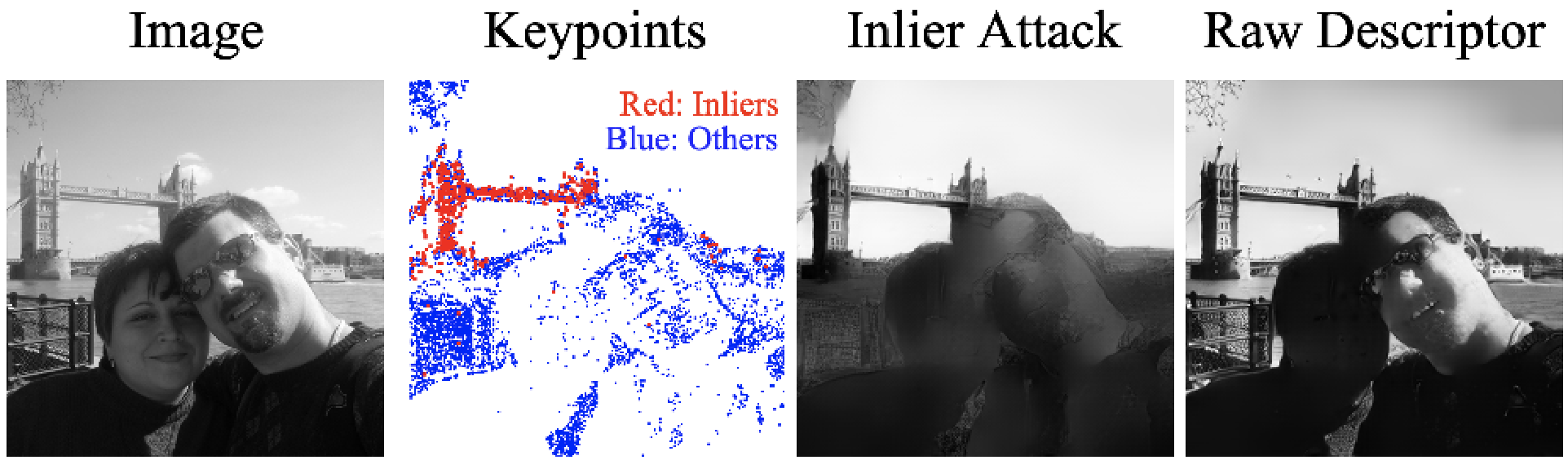}
\caption{\textbf{Inlier Attack.}}
\label{fig:attack_inlier}
\end{figure}

\subsection{Sensitivity of Inlier Content}
Since inlier correspondences emerge from RANSAC in the geometric tasks, one natural attack one may think of is leveraging these inlier features to reveal the image content; we term this as inlier attack. 
We note that this attack is generally applicable to all privacy protocols that are capable of geometric utility tasks where RANSAC returns inliers, \eg ours and \cite{dusmanu2021privacy}. However, RANSAC inliers typically consist of only static background scenes without dynamic foreground (e.g. faces). We clarify that the privacy protection mainly targets at the foreground in both our and \cite{dusmanu2021privacy}'s problem setup, and thus the inlier attack was not a concern. Nonetheless, we perform inlier attack here and show example result in \ref{fig:attack_inlier}. As expected, the attack works only for the background bridge, but not for the foreground faces.

\end{document}